\documentclass[10pt,twocolumn,letterpaper]{article}

\usepackage{iccv}
\usepackage{times}
\usepackage{epsfig}
\usepackage{graphicx}
\usepackage{amsmath}
\usepackage{amssymb}
\usepackage{dsfont}
\usepackage{amssymb}
\usepackage{tipa}
\usepackage{mathrsfs}
\usepackage{textcomp}
\usepackage{tabularx}
\usepackage{booktabs}
\graphicspath{{./images/}}
\DeclareMathOperator*{\argmin}{argmin}

\usepackage[pagebackref=true,breaklinks=true,letterpaper=true,colorlinks,bookmarks=false]{hyperref}

 \iccvfinalcopy 

\def\iccvPaperID{194} 

\ificcvfinal\fi
\begin{document}

\title{Synthetic to Real Adaptation with \\ Generative Correlation Alignment Networks}
\author{Xingchao Peng\\
Boston University\\
{\tt\small xpeng@bu.edu}
\and
Kate Saenko\\
Boston University\\
{\tt\small saenko@bu.edu}
}

\maketitle

\begin{abstract}
Synthetic images rendered from 3D CAD models are useful for augmenting training data for object recognition algorithms. However, the generated images are non-photorealistic and do not match real image statistics. This leads to a large domain discrepancy, causing models trained on synthetic data to perform poorly on real domains. Recent work has shown the great potential of deep convolutional neural networks to generate realistic images, but has not utilized  generative models to address synthetic-to-real domain adaptation. In this work, we propose a Deep Generative Correlation Alignment Network (DGCAN) to synthesize images using a novel domain adaption algorithm. DGCAN leverages a shape preserving loss and a low level statistic matching loss to minimize the domain discrepancy between synthetic and real images in deep feature space. Experimentally, we show training off-the-shelf classifiers on the newly generated data can significantly boost performance when testing on the real image domains (PASCAL VOC 2007 benchmark and Office dataset), improving upon several existing methods. 
\end{abstract}


\vspace{-0.3cm}
\section{Introduction}
\label{intro}

\begin{figure}[tp]
\centering
\includegraphics[width=\linewidth]{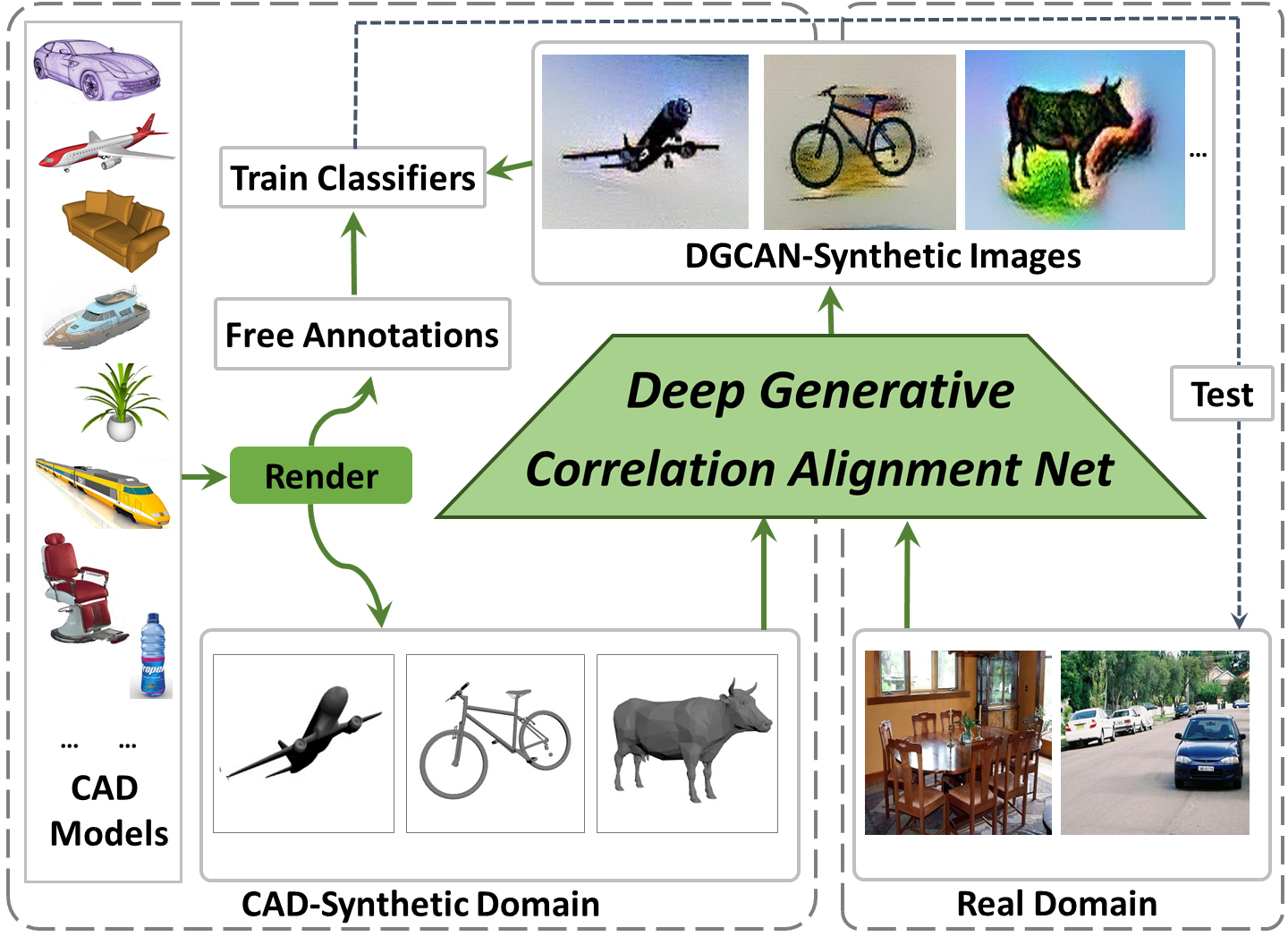}
\caption{\small {\bf
Overview of our approach}
We propose a Deep Generative Correlation Alignment Network (\textit{DGCAN}) to bridge the domain gap between CAD-synthetic and real images in deep feature space.  \textit{DGCAN} can generate inexpensive annotated training data by blending the object shape from freely available 3D CAD models together with structured texture from a small amount of real background images. We train off-the-shelf classifiers on the DGCAN-synthetic images and test them on the real image domain, demonstrating a significant improvement over existing adaptation methods.  
}
\label{fig_overview}
\vspace{-0.2cm}
\end{figure}

Recent advances achieved by Deep Convolutional Neural Networks (DCNN) ~\cite{RCNN, alexnet, su2015render, sun2015return, vgg, he2015deep} are unfortunately hampered by their  dependence on massive amounts of training examples. Ad-hoc collection and annotation of training data for various computer vision applications is cumbersome and costly. 3D CAD simulation is a promising solution to this problem~\cite{peng2015learning, liebelt2010multi, stark2010back, su2015render, sun2009multi}. Rendering images from freely available CAD models can potentially produce an infinite number of training examples from many viewpoints and for almost any object category. Previous work~\cite{peng2015learning} utilized computer graphics (CG) technique to render 2D CAD-synthetic images and consequently train deep CNN-based classifiers on them. However, their CAD-synthetic images are highly non-realistic due to the absence of natural object texture and background. More specifically, they exhibit the following problems: 1) large mismatch between foreground and background, 2) higher contrast between the object edges and the background, 3) non-photorealistic scenery. These problems inevitably lead to a significant domain shift between CAD-synthetic and real images.

To minimize the domain shift, domain adaptation methods have been proposed to align two domains in manifold space~\cite{gong2012geodesic, fernando-iccv13} or in deep feature space~\cite{long2015learning, long2016unsupervised, SunS16a}. These algorithms bridge the domain gap between real-image domains. However, the domain shift between synthetic and real domains is much larger than that between two real-image domains. CAD-to-real adaptation methods~\cite{massa2016deep, Dosovitskiy_2015_CVPR} have been proposed but they only align the viewpoint of specific indoor categories and cannot be directly applied to recognition systems in the wild. 

Our main idea is to incorporate domain adaptation algorithm into generative networks. Generative neural networks have recently been proposed to create novel imagery that shares common properties with some given images, such as \textit{content} and \textit{style}~\cite{gatys2015neural}, similarity in feature space~\cite{cogan, goodfellow2014generative, viscnn, mahendran14arXiv, mirza2014conditional}, etc.
However, these approaches have several limitations for use in domain adaptation. For example, \textit{Generative Adversarial Nets} (GANs)~\cite{goodfellow2014generative} and \textit{style transfer} approaches~\cite{gatys2015neural, gatys2016preserving} can generate images but are not designed for domain adaptation. \textit{Coupled GANs}~\cite{cogan} only handle domain shifts between small images (28$\times$28 pixel resolution).  \textit{Conditional GANs}~\cite{IsolaZZE16} can learn image-to-image translation but need paired training data that are costly to obtain, \ie CAD models and corresponding natural images.

To overcome the limitations of the aforementioned approaches, we propose a \textit{Deep Generative Correlation Alignment Network} (\textit{DGCAN}) to bridge the domain discrepancy between CAD-synthetic and real images. Our work is primarily motivated by~\cite{peng2015learning,gatys2015neural,SunS16a}. As shown in Figure~\ref{fig_overview}, we generate novel images by matching the convolutional layer features with those of a \emph{content} CAD-synthetic image and the feature statistics of a real image containing a background scene. Unlike \textit{neural style}~\cite{gatys2015neural}, the goal is not to create an artistic effect but rather to adapt the CAD-synthetic data to match the statistics of real images and thus improve generalization. To this end, we employ the correlation alignment (CORAL) loss~\cite{SunS16a} for adaptation. However, instead of learning to align features, we generate images whose feature correlations match the target real-image domain.

Our synthesized results reveal that \emph{DGCAN} can satisfactorily blend the contour of specific objects (from CAD-synthetic images) with natural textures from real images. Although the generated images are not fully photorealistic, they appear to have more natural statistics to the deep network, improving its performance. Extensive experiments on the PASCAL and Office dataset show that our approach yields a significant performance boost compared to the previous state-of-the-art methods~\cite{peng2015learning,sun2015return,SunS16a, fernando-iccv13, gong2012geodesic, long2015learning, long2016unsupervised, gatys2015neural}.

The contributions of this paper can be summarized as follows.
\begin{itemize}
    \item We propose \textit{Deep Generative Correlation Alignment Network} (\textit{ DGCAN}) to synthesize CAD objects contour from the CAD-synthetic domain with natural textures from the real image domain. 
    \item We explore the effect of applying the content
    and \emph{CORAL} losses to different layers and determine the optimal configuration to generate the most promising stimuli.
    \item We empirically show the effectiveness of our model over several state-of-the-art methods by testing on real image datasets.
\end{itemize}

\section{Related Work}
\label{related}

\begin{figure*}[t]
\centering
\includegraphics[width=\linewidth]{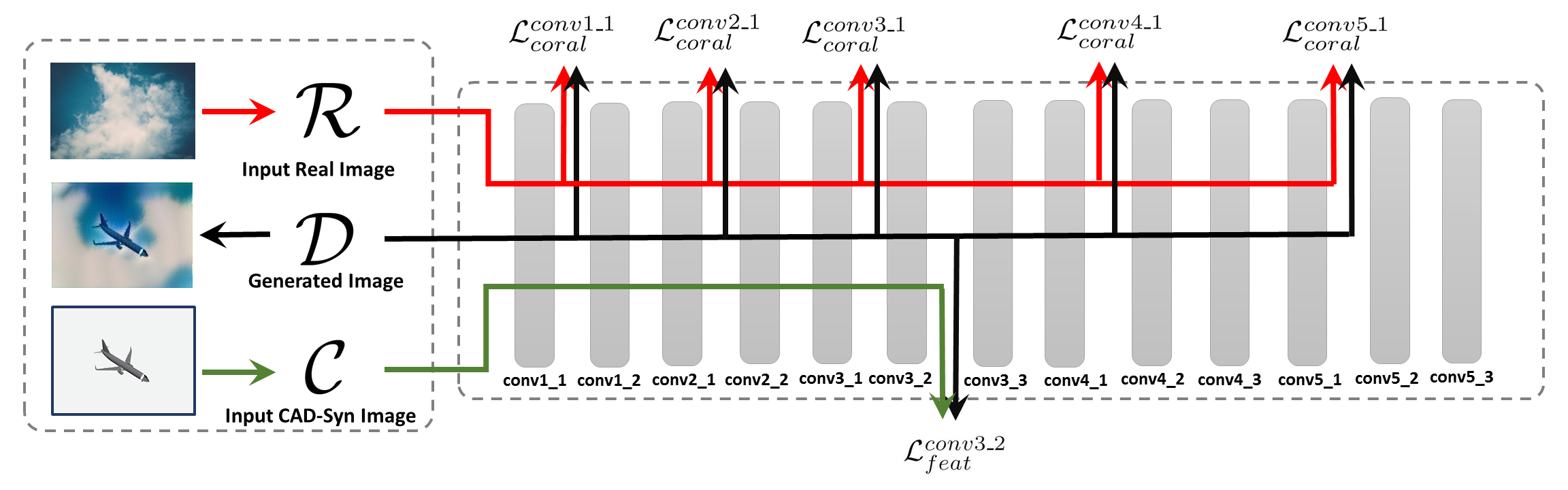}
\caption{{\bf Illustration of \emph{DGCAN}.}  Our model, the Deep Generative Correlation Alignment Network (DGCAN), takes CAD-synthetic images $\mathcal{C}$ and real background images $\mathcal{R}$ as input and generates  novel images $\mathcal{D}$ that contain the same object shapes as the CAD images but have more natural feature distributions by transferring the texture from the background image. The network structure is based on VGG-16~\cite{vgg}, which comprises 13 convolutional layers, divided in five groups. The image is generated by applying the $\ell^2$ loss $\mathcal{L}^{\mathcal{X}_f}_{feat}$ and \emph{CORAL} loss $\mathcal{L}^{\mathcal{X}_c}_{coral}$ to the following layers: $\mathcal{X}_f=\{conv3\_2\}$, $\mathcal{X}_c=\{conv1\_1, \,  conv2\_1, \, conv3\_1, \, conv4\_1, \, conv5\_1   \}$. 
}
\label{fig_dcon}
\end{figure*}

\vspace{0.1cm}
\noindent \textbf{CAD Simulation} CAD simulation has been extensively used by researchers since the early days of computer vision~\cite{nevatia77}. 3D CAD models have been utilized  to generate stationary synthetic images with variable object poses, textures, and backgrounds~\cite{peng2015learning}. Recent usage of CAD simulation has been extended to multiple vision tasks, e.g. object detection~\cite{peng2015learning,massa2016deep}, pose estimation~\cite{liebelt2010multi, stark2010back, su2015render, sun2009multi}, robotic simulation~\cite{tzeng2015towards}, semantic segmentation~\cite{richter2016playing}. However, for many tasks, CAD-synthetic images are too low-quality due to the absence of realistic backgrounds and texture. To mitigate this drawback,~\cite{peng2015learning} proposes to directly add auxiliary texture and background to the rendered results, with the help of commercial software (e.g. AutoDesk 3ds MAX\footnote{http://www.autodesk.com/store/products/3ds-max}). However, this method  introduces new problems, such as unnatural positioning of objects (e.g. floating car above the road), high contrast between object boundaries and background, etc. Our approach tackles these problems by synthesizing novel imagery with \emph{DGCAN} and can generate images with natural feature statistics. 

\vspace{0.1cm}
\noindent \textbf{DCNN Image Synthesis} Deep convolutional neural networks learn distributed, invariant and nonlinear feature representations from large-scale image repositories~\cite{bengio2013representation}. \textit{Generative Adversarial Networks} (GANs)~\cite{goodfellow2014generative} and their variations~\cite{mirza2014conditional, nguyen2016synthesizing} aim to synthesize images that are indistinguishable from the distribution of images in their training set. However, training GANs is difficult and often leads to oscillatory behavior. \textit{Style transfer}~\cite{gatys2015neural}~synthesizes novel stimuli by aligning the \emph{conv} layer features and \emph{Gram Matrices} of the features. In this way, the synthesized image simultaneously preserves the arrangement of a content image (often a normal photograph) and the colours and subtle local structures of a style image (often an artist's work). Our approach is inspired by \textit{style transfer}~\cite{gatys2015neural}, but is geared towards adapting a set of CAD-synthetic images to real image domain with a domain adaptation loss.

 \vspace{0.1cm}
 \noindent \textbf{Domain Adaptation} Domain shift results in a significant performance degradation when recognition systems are trained on one domain (source) and tested on another (target). Shallow domain adaptation algorithms aim to bridge the two feature distributions via mappings learned either by minimizing a distribution distance metric~\cite{borgwardt2006integrating,sun2015return}, or by projecting the feature distributions to a common low-dimensional manifold~\cite{gopalan-iccv11, gong2012geodesic, long2014transfer}. Deep domain adaptation methods address the domain shift by adding one or multiple adaptation layers and losses~\cite{tzeng2014deep, tzeng2015towards, long2015learning, SunS16a}, or use an adversarial network to match the source distribution to target~\cite{tzeng2015towards, cogan}. All of the aforementioned methods follow the paradigm of aligning the source domain and target domain in feature space. In contrast, we take a generative approach to combine the statistics of target domain images with the content of source domain images. Recent proposed generative models~\cite{bousmalis2016unsupervised, cogan} adapt two domains by adversarial losses. However, these methods only generate small images. Our model can generate large images with arbitrary resolution.

\section{Approach}
\label{approach}

Suppose we are given $n_s$ labeled source-domain CAD-synthetic (image, label) pairs $\mathcal{I}_s = \{\mathcal{C}_i, \mathcal{Y}_i\}^{n_s}_{i=1}$, and $n_t$ target-domain real images $\mathcal{I}_t = \{\mathcal{R}_i\}^{n_t}_{i=1}$.
We assume that the target domain is unlabeled, so object classifiers can only be trained on $\mathcal{I}_s$.  However, their performance will degrade when testing on $\mathcal{I}_t$ due to the domain discrepancy. Our aim is to synthesize a labeled intermediate dataset $\mathcal{I} = \{\mathcal{D}_i, \mathcal{Y}_i\}^{n}_{i=1}$, such that each $\mathcal{D}_i \in \mathcal{I}$ contains a similar object shape and contour with $\mathcal{C}_i \in \mathcal{I}_s$ and similar local pattern, color, and subtle structure (``style'' as illustrated in~\cite{gatys2015neural}) with some random $\mathcal{R} \in \mathcal{I}_t$.

To generate $\mathcal{D}$ from $\mathcal{C}$ and $\mathcal{R}$, the most straightforward method is to average the two images. Traditional computer vision blending approaches, such as half-half alpha blending or pyramid blending lead to image artifacts that contribute to the domain shift. Previous CG-based method~\cite{peng2015learning} applied real image background and texture to CAD models, leading to the problems illustrated in Section~\ref{intro}. 
Instead, we propose to align the generated $\mathcal{D}$ to $\mathcal{C}$ and $\mathcal{R}$ in the DCNN feature space, as shown in Figure~\ref{fig_dcon}. Analogously to~\cite{gatys2015neural}, our model synthesizes an image $\mathcal{D}$ from $\mathcal{C}$ with $\mathcal{D} \sim p(\mathcal{D} | \mathcal{C}, \mathcal{R})$. The generation is guided by two losses, one to ensure the object contour stays the same, and the other to ensure the image has similar low-level statistics with real images.

\subsection{Deep Convolutional Neural Network}
We base our approach on the VGG-16~\cite{vgg} network which consists of 13 convolutional layers (\emph{conv1\_1-conv5\_3}), 3 fully connected layers (\emph{fc6-fc8}) and 5 pooling layers (\emph{pool1-pool5}). 
The convolutional layers consist of a set of learnable kernels. Each kernel is convolved with the input volume to compute hidden activations during the forward pass and its parameters are updated through a back-propagation pass. 
We denote $\mathcal{H}^l(\cdot)$ as DCNN's $l^{th}$ layer's representation matrix, $\mathcal{H}^l_i(\cdot)$ as the $i^{th}$ dimension of $\mathcal{H}^l(\cdot)$ and $\mathcal{H}^l_{ij}(\cdot)$ as $j^{th}$ value of $\mathcal{H}^l_i(\cdot)$.

\subsection{Shape Preserving loss}
To preserve the shape information of CAD-synthetic images, we propose to use the $\ell^2$ loss in feature space as follows
\begin{equation}
\label{equ_f_loss}
    \mathcal{L}^{\mathcal{X}_f}_{feat} = \sum_{l \in \mathcal{X}_f}(\frac{\omega_{f}^l}{2\alpha^l}\sum_{i}\lVert\mathcal{H}_i^l(\mathcal{D}) - \mathcal{H}_i^l(\mathcal{C}) \rVert^2_2).
\end{equation}

where $\mathcal{D} \in \mathcal{I}$, $\mathcal{C} \in \mathcal{I}_{s}$; $\omega_{f}^l$ is the loss weight of $l^{th}$ layer in DCNN feature space; $\mathcal{X}_f$ is the collection of convolutional layers which the $\ell^2$ loss is applied to; $\alpha^l = N^lF^l$, where $N^l$ is the channel number of $l^{th}$ layer's feature, and $F^l$ is the length of feature in a single channel. 

The derivative of this loss with respect to the activations in a particular layer $l$ can be computed by:
\begin{equation}
\label{equ_f_deriv}
    \frac{\partial\mathcal{L}^{\mathcal{X}_f}_{feat}}{\partial\mathcal{H}_{ij}^l(\mathcal{D})} = \frac{\omega_{f}^l}{\alpha^l}(\mathcal{H}_{ij}^l(\mathcal{D}) - \mathcal{H}_{ij}^l(\mathcal{C}))
\end{equation}

This gradient can be back-propagated to update the pixels while synthesizing $\mathcal{D}$.

\begin{figure*}
    \centering
    
    \includegraphics[width=\textwidth]{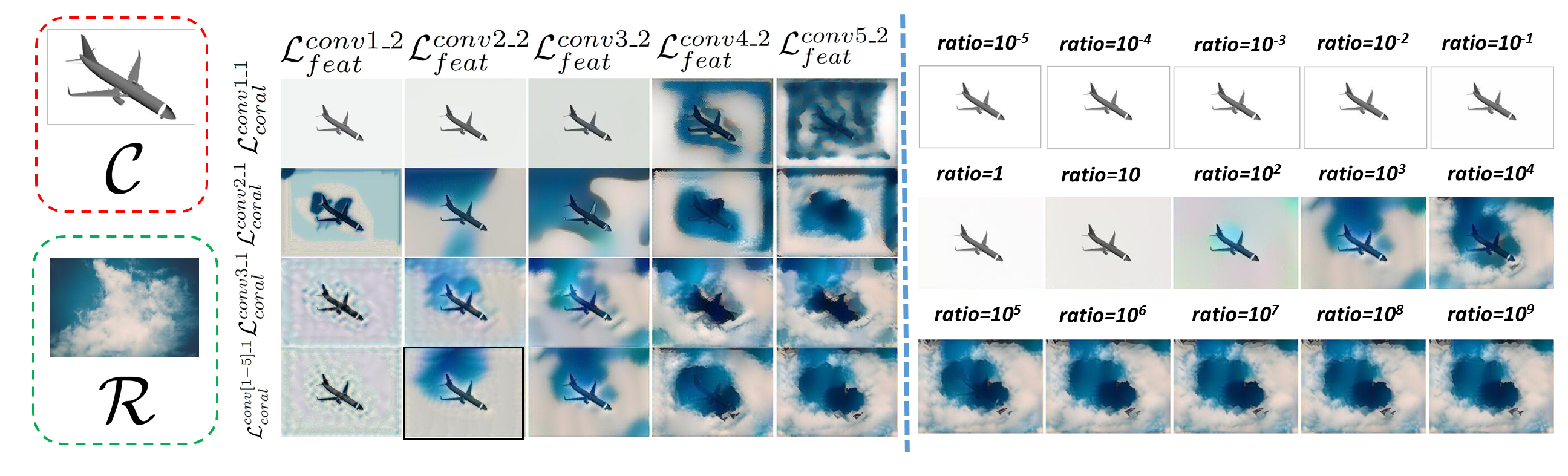}
  \caption{{\bf Illustration of our synthesized results.} We leverage \emph{DGCAN} to synthesize novel images based on two inputs, i.e. source domain CAD-synthetic image $\mathcal{C} \in \mathcal{I}_s$ and target domain real background image $\mathcal{R} \in \mathcal{I}_t$. \textbf{(1).} We exhaustively apply $\mathcal{L}_{feat}$ and $\mathcal{L}_{coral}$ to different \emph{conv} layers to find the best configuration. The results (\textbf{left plot}) demonstrate that \emph{DGCAN} can generate more distinct object contours when applying $\mathcal{L}_{feat}$ to lower \emph{conv} layers and can synthesize more structured style texture when applying the $\mathcal{L}_{coral}$ to higher \emph{conv} layers. ($\mathcal{L}^{[1-5]\_1}_{coral}$ means applying $\mathcal{L}_{coral}$ to $conv1\_1,\;conv2\_1,\;conv3\_1,\;conv4\_1,\;conv5\_1$ simultaneously) \textbf{(2).} We vary the trade-off parameter $\lambda$ in equation~\ref{equ_dcn} from $10^{-5} \sim 10^{9}$ to learn the optimal value for $\lambda$. The results (\textbf{right plot}) show that the shape contour  dominates the background texture when $\lambda$ is small.}
  \label{fig_res} 

\end{figure*}

\subsection{Naturalness loss}
Networks trained on CAD images will not work well on input real images because of the mismatch in low-level image statistics such as textures, edge contrast, color, etc. To align the low-level texture statistics of the generated images to the real image domain, we propose to employ the CORAL loss.
Correlation Alignment (CORAL) was first devised by~\cite{sun2015return} to match the second-order statistics of feature distributions for domain adaptation. It is derived by minimizing the domain discrepancy with squared \emph{Frobenius} norm $min\lVert Cov_{S}-Cov_{T}\rVert_{F}^2$, where $Cov_{S}, Cov_{T}$ are the covariance matrices of feature vectors from source domain and target domain, respectively. This problem is equivalent to solving $A^{\star} = \argmin \lVert A^{\top}Cov_{S}A - Cov_{T} \rVert^2_{F}$. 

Inspired by~\cite{sun2015return}, we define the CORAL loss $\mathcal{L}^{\mathcal{X}_c}_{coral}$ as
\begin{equation}
\label{equ_c_loss}
    \mathcal{L}^{\mathcal{X}_c}_{coral} = \sum_{l \in \mathcal{X}_c}(\frac{\omega^l_{c}}{4{\alpha^l}^2} \lVert Cov(\mathcal{H}^l(\mathcal{D})) - Cov(\mathcal{H}^l({\mathcal{R}})) \rVert^2_F)
\end{equation}

where $\mathcal{D} \in \mathcal{I}$, $\mathcal{R} \in \mathcal{I}_{t}$; $\omega_{c}^l$ is the $CORAL$ loss weight of $l^{th}$ layer; $\mathcal{X}_c$ is the collection of convolutional layers that the $CORAL$ loss is applied to; $Cov(\cdot)$ is the covariance matrix of $l^{th}$ layer's activation; $\lVert \cdot \rVert_F$ denotes the \emph{Frobenius} distance.

Analogous to~\cite{SunS16a}, the covariance matrices are given by:
\begin{equation}
\begin{split}
     Cov(\mathcal{H}^l(\mathcal{M})) = &\frac{1}{N^l}\{{\mathcal{H}^l(\mathcal{M})}^\top\mathcal{H}^l(\mathcal{M}) \\ &- \frac{1}{N^l}[(\textbf{1}^{\top}\mathcal{H}^l(\mathcal{M}))^{\top}(\textbf{1}^{\top}\mathcal{H}^l(\mathcal{M}))]\}
\end{split}
\end{equation}
where $\mathcal{M} \in \{\mathcal{D}, \mathcal{R}\}$, \textbf{1} is a column all-one vector, and $N^l$ is the number of feature channels in $l^{th}$ layer.

The derivative of the \emph{CORAL} loss with respect to a particular layer $l$ can be calculated with chain rule:
\begin{equation}
\label{equ_c_deriv}
\begin{split}
    \frac{\partial \mathcal{L}^{\mathcal{X}_c}_{coral}}{\partial \mathcal{H}_{ij}^l(\mathcal{D})} = 
    &   \frac{\omega^l_{c}}{N^l{\alpha^l}^2} \{[{\mathcal{H}^l(\mathcal{D})}^{\top}-\frac{1}{N^l}(\textbf{1}^{\top}\mathcal{H}^l(\mathcal{D}))\textbf{1}^{\top}]^{\top} \\ &\cdot(Cov^l(\mathcal{D}) - Cov^l(\mathcal{R}))\}_{ij}
    \end{split}
\end{equation}

Our final method combines the loss functions defined by equation~\ref{equ_f_loss} and equation~\ref{equ_c_loss}. We start from an image $\mathcal{D} \in \mathcal{I}_s$ and pre-process it by adding a random perturbation $\epsilon$, where $\epsilon \sim \mathcal{N}(0, \Sigma)$. We then feed the image forward through DGCAN and compute the $\ell^2$ loss with respect to $\mathcal{D}$ and CORAL loss with respect to $\mathcal{R}$. The back-propagated gradient thus guides the  image synthesis process. Hence, the synthesized image is the output of the function:
\begin{equation}
\label{equ_dcn}
\mathcal{D}^{\star} = \underset{\mathcal{D} \in \mathcal{I}} \argmin ( \mathcal{L}^{\mathcal{X}_f}_{feat}+\lambda \mathcal{L}^{\mathcal{X}_c}_{coral}|\mathcal{C}, \mathcal{R},\mathcal{X}_f,\mathcal{X}_c, \lambda,\epsilon)
\end{equation}

where $\mathcal{L}^{\mathcal{X}_f}_{feat}+\lambda \mathcal{L}^{\mathcal{X}_c}_{coral}$ denotes the total loss of \textit{DGCAN}, and $\lambda$ denotes the trade-off weight between $\ell^2$ loss and the \emph{CORAL} loss. The hyperparameter $\lambda$ is set through cross validation. 

\begin{figure*}[t]
    \centering
    
    \includegraphics[width=\textwidth]{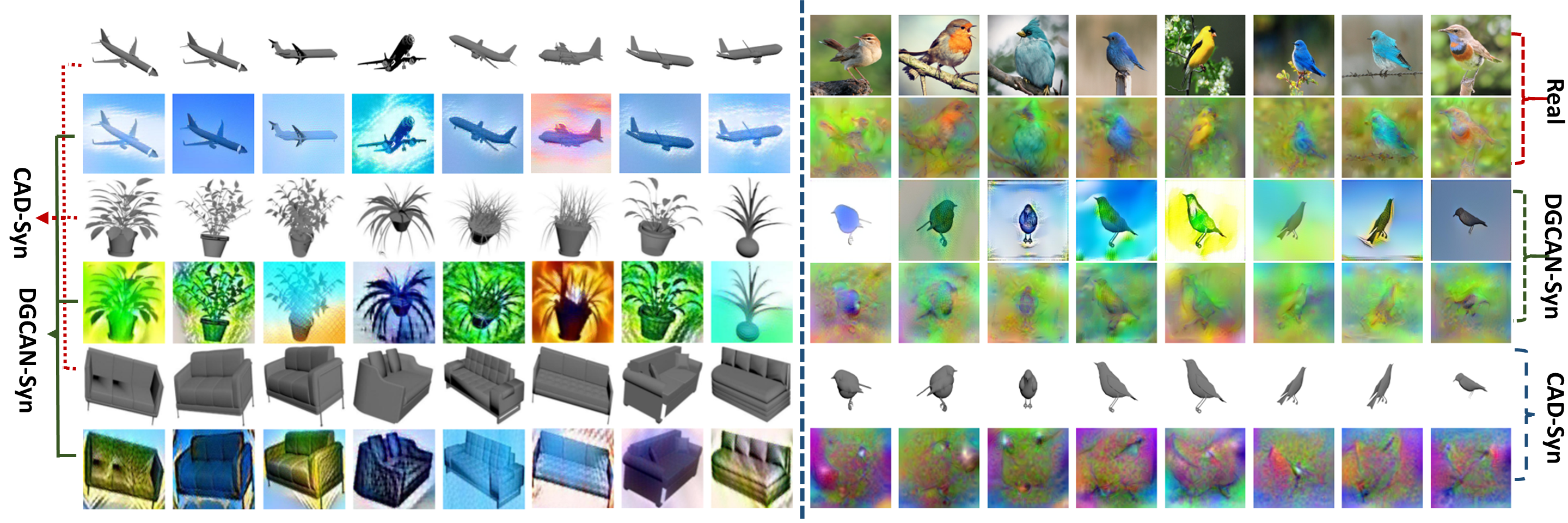}
    \caption{\textbf{DGCAN-synthetic examples and visualizations.} \textbf{(1)} The left plot shows randomly selected DGCAN-synthetic examples ($\mathcal{D}$) and their corresponding CAD-synthetic images ($\mathcal{C}$). The rendered results demonstrate that \emph{DGCAN} can synthesize novel images with clear object contours and photo-realistic textures.\textbf{ (2)} The right plot illustrates the reconstruction results generated by using the tools provided by~\cite{mahendran2016visualizing}. The reconstructions reveal that our DGCAN-synthetic images share more similarities with real images from the DCNN's perspective. The (uniform gray-scale) CAD-synthetic images only provides edge information. Thus, the pixels in the reconstructed images are dominated by the rich color and texture information encoded in the DCNN's parameters. (Best viewed in color!)}
    \label{fig_plotres}
    \vspace{-0.4cm}
\end{figure*}  
\section{Experiments}
\label{sec_exp}

Our experiments include two parts. First, we apply \emph{DGCAN} to the CAD-synthetic dataset provided by~\cite{peng2015learning} to synthesize adapted DGCAN-synthetic images. Second, we train off-the-shelf classifiers on the DGCAN-synthetic images and test on the PASCAL 2007~\cite{pascal-voc-2007} and Office~\cite{saenko2010adapting} datasets. We implement our model with the  \textit{Caffe}~\cite{caffe} framework. Datasets (both CAD-synthetic and DGCAN-synthetic), code and experimental configurations will be made available publicly.

\subsection{Generating Adapted Images}
As shown in Figure~\ref{fig_dcon}, while generating the DGCAN-synthetic dataset, we set CAD-synthetic images as $\mathcal{C}$ and real images downloaded from the Google image search engine as $\mathcal{R}$.

\vspace{0.05cm}
\noindent \textbf{CAD-Synthetic Dataset} The CAD-synthetic dataset in~\cite{peng2015learning} was rendered from 3D CAD models for zero-shot or few-shot learning tasks. The dataset contains 6 subsets with different configurations (i.e. RR-RR, W-RR, W-UG, RR-UG, RG-UG, RG-RR).
The process of rendering the dataset (we refer the reader to~\cite{peng2015learning} for more details) can be summarized as follows: (1) collecting 3D-CAD models from large-scale on-line repositories (Google Sketchup, Stanford 3D ShapeNet\footnote{http://shapenet.cs.stanford.edu/}), (2) selecting image cues (background, texture, pose, etc.), (3) rendering synthetic images with AutoDesk 3ds Max. In our experiments, we only adopt images with white background because other subsets suffer from the issues described in Section~\ref{intro}.

\vspace{0.05cm}
\noindent \textbf{Parameter tuning} To determine the optimal configuration for $\mathcal{X}_f$, $\mathcal{X}_c$ and $\lambda$, we exhaustively apply $\mathcal{L}_{feat}$, $\mathcal{L}_{coral}$ to different \emph{conv} layers and vary $\lambda$ from $10^{-5} \sim 10^{9}$ on a small validation dataset.

\vspace{0.05cm}
\noindent \textbf{Results and Analysis} A representative subset of rendered results with different setting for $\mathcal{X}_f$, $\mathcal{X}_c$ and $\lambda$ are shown in Figure \ref{fig_res}. The left plot shows the effect of different configurations of $\mathcal{X}_f$ and $\mathcal{X}_c$. The results demonstrate that when $\mathcal{L}_{feat}$ is applied to lower \emph{conv} layers, \emph{DGCAN} can generate more distinct contour of the object from CAD-synthetic data and when $\mathcal{L}_{coral}$ is applied to higher \emph{conv} layers, \emph{DGCAN} can generate more structured texture. Empirical evidence~\cite{gatys2015neural} shows that this effect mainly stems from two factors. ason is the increasing receptive field size, given the receptive field sizes of VGG-16's \emph{conv1\_2, conv2\_2, conv3\_2, conv4\_2, conv5\_2} are 5, 14, 32, 76 and 164, respectively. The second factor is the increasing feature complexity along the network hierarchy.

\begin{table*}[t]
\centering
\scriptsize
\noindent\begin{tabular}{ p{2.1cm} || p{0.27cm}  p{0.27cm}  p{0.27cm}  p{0.27cm} p{0.27cm} p{0.27cm} p{0.27cm} p{0.27cm} p{0.27cm} p{0.27cm} p{0.27cm} p{0.27cm} p{0.27cm} p{0.27cm} p{0.27cm} p{0.27cm} p{0.27cm} p{0.27cm} p{0.27cm} p{0.27cm} |p{0.6cm}    }
\hline
 Method  & aero & bike & bird & boat & bottle & bus & car & cat & chair & cow & table & dog & horse & mbike & person & plant & sheep & sofa & train & tv & overall\\
 
\hline
CAD-AlexNet~\cite{peng2015learning}& 25.7 &19.8 &11.8 &31.3 &11.4 &72.0 &26.0&8.4 &12.2 &27.7 &1.0 &3.2 &3.3 &69.1 &11.6 &40.0 &0.0 &15.4 &43.7 &40.4 &18.48\\
CORAL-Alex~\cite{sun2015return}& 33.1 &15.9 &17.7 &27.7 &3.5 &79.9 &26.5 &17.6 &15.0 &22.8 &5.7 &10.0 &11.1 &62.1 &12.3 &29.1 &0.0 &9.1 &25.5 &26.6 & 18.18 \\
DCORAL-Alex~\cite{SunS16a}&28.0 &28.5 &9.5 &26.5 &25.4 &65.7 &41.3 &21.6 &22.7 &52.0 &1.0 &5.8 &13.4 &71.8 &22.0 &33.3 &3.9 &9.6 &34.1 &60.7 & 24.48\\
SA-fc7~\cite{fernando-iccv13} & 59.8 &46.5 &34.2 &29.3 &6.2 &66.9 &28.4 &31.4 &15.6 &23.1 &9.7 &11.7 &19.7 &63.7 &10.3 &29.9 &9.0 &24.5 &16.2 &46.5 &21.10 \\
GFK-fc7~\cite{gong2012geodesic} &43.4 &31.4 &18.6 &41.7 &4.6 &65.7 &24.5 &10.3 &16.8 &14.9 &15.1 &4.9 &18.0 &50.7 &5.2 &23.1 &12.9 &17.2 &14.9 &29.4 &16.14 \\
DAN~\cite{long2015learning}&35.7 &55.8 &23.6 &19.8 &17.7 &73.6 &43.1 &10.8 &31.5 &58.7 &2.0 &4.9 &7.1 &63.4 &12.4 &44.9 &0.3 &15.7 &13.6 &28.0 &23.97\\
RTN~\cite{long2016unsupervised}&20.9 &3.6 &2.6 &12.0 &1.5 &69.7 &37.2 &33.8 &21.8 &53.2 &0.0 &0.2 &0.3 &61.0 &11.6 &21.5 &0.0 &3.5 &7.6 &44.6 &17.76 \\
StyleTransfer\cite{gatys2015neural}&59.5 &61.2 &16.7 &31.6 &47.6 &39.8 &24.1 &25.7 &45.2 &35.0 &5.0 &10.0 &18.0 &35.0 &14.5 &39.7 &2.6 &4.0 &9.9 &47.9 &24.78 \\
\textbf{DGCAN-AlexNet}& 62.4 &60.7 &20.3 &13.2 &16.3 &65.4 &7.2 &33.2 &43.0 &31.6 &4.0 &3.4 &15.7 &48.0 &26.8 &51.5 &0.6 &14.1 &10.3 &69.5 &\textbf{27.46}\\
\hline
\end{tabular}
\vspace{0.05cm}

\caption{\textbf{Results on PASCAL 2007}.
We show per-category accuracy of different methods based on ``AlexNet''~\cite{alexnet} features. 
The results clearly demonstrate the superiority of our model over CAD-synthetic method~\cite{peng2015learning}, \textit{style transfer} method~\cite{gatys2015neural} and several state-of-the-art domain adaptation models~\cite{sun2015return, SunS16a, fernando-iccv13, gong2012geodesic, long2015learning, long2016unsupervised}}

 \label{tab_cls}
\end{table*}

\begin{figure*}[t]
    \centering
    \includegraphics[width=\textwidth]{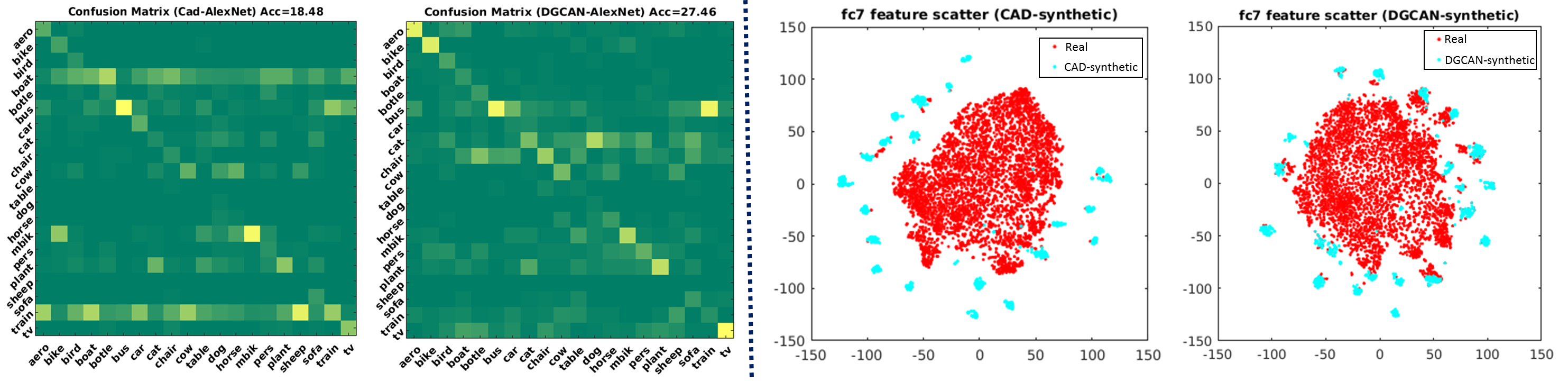}
    \caption{\textbf{Confusion Matrices and t-SNE plots of \textit{fc7} feature.} \textbf{(1).} The confusion matrices on the left show models trained on DGCAN-synthetic dataset (right subplot) pose a different error mode from those trained on CAD-synthetic dataset (left subplot). \textbf{(2)} The t-SNE plots on the right shows the embedded \textit{fc7} features of realistic and synthetic images are better aligned after applying our model to CAD-synthetic images. (Best viewed in color!) }
    \label{fig_conf}
\end{figure*}

\begin{table}[t]
\centering
\noindent\begin{tabular}{p{2.5cm} p{1.3cm} p{1.3cm} p{1.3cm}}
\toprule
&VGG&ResNet&AlexNet\\
\midrule
CAD~\cite{peng2015learning} & 10.30 & 13.13 & 18.48\\
CORAL~\cite{sun2015return} & 11.67 & 12.24 & 18.18\\
DCORAL~\cite{SunS16a} &17.76 & - & 24.48\\
SA~\cite{fernando-iccv13} &20.38 & 19.33 & 21.10\\
GFK~\cite{gong2012geodesic} &17.05 & 18.43 &16.14\\
\\
\textbf{DGCAN} &\textbf{22.92} & \textbf{20.59} & \textbf{27.46}\\
\bottomrule
\end{tabular}
\vspace{0.05cm}
\caption{\textbf{Results on PASCAL 2007}. We train three off-the-shelf classifiers on the adapted dataset and test on PASCAL 2007 benchmark. The results demonstrate the our model works better than other domain adaptation methods~\cite{SunS16a, sun2015return, fernando-iccv13, gong2012geodesic} with all the three classifiers.}
\label{tab_cls_vgg_res}
\end{table}

To find the optimal trade off ratio $\lambda$, we synthesize images with $\lambda$ ranging from $10^{-5}$ to $10^{9}$. The right plot in Figure~\ref{fig_res} reveals when $\lambda$ ($\mathcal{L}_{coral}$ to $\mathcal{L}_{feat}$ ratio) is small, the object contour will dominate the background texture cues. On the contrary, when $\lambda$ is increased to $10^5$, the contour of the object gradually fades away and more structured textures from the real image emerge.

We randomly select some rendered results from three categories (``aeroplane'', ``potted plant'', ``sofa'') and show them in the left plot of Figure~\ref{fig_plotres}. The images are generated with the configuration $\mathcal{X}^f = {conv3\_2}$, $\mathcal{X}^c = {conv[1-5]\_1}$ ($\omega^{1\sim5}_{c}=0.2$) and $\lambda=10^3$. The results demonstrate that DGCAN-synthetic imagery preserves clear object contours from the CAD-synthetic images and synthesized textures from realistic domain.

We further leverage the DCNN visualization tool provided by~\cite{mahendran2016visualizing} to show that DGCAN-synthetic images share more similarities with real images.~\cite{mahendran2016visualizing} provides an effective tool to reconstruct  an image from its representation. We compare the reconstruction results of bird images from three domains, \ie DGCAN-synthetic, CAD-synthetic and real domains. In the right subplot of Figure~\ref{fig_plotres}, the odd rows show the original bird images, and their corresponding reconstructions are located in the even rows. From the plots, we can observe recognizable bird shapes from the reconstructed images of DGCAN-synthetic images. However, the birds in the reconstructed images of CAD-synthetic domain are lost in noisy color patches. These visualization results demonstrate that the DCNN can better recover category information from DGCAN-synthetic images than from CAD-synthetic images. 


\subsection{Domain Adaptation Experiments}

In this section, we evaluate our approach on CAD-to-real domain adaptation tasks, using object classification as the application. The goal is to generate adapted CAD images using our approach, then train deep object classifiers on the data, and test on real-image benchmarks. We compare the effectiveness of our model to previous methods~\cite{alexnet, SunS16a, sun2015return, gong2012geodesic, long2015learning, long2016unsupervised, gatys2015neural, fernando-iccv13} on two benchmarks: PASCAL VOC 2007~\cite{pascal-voc-2007} and the Office~\cite{saenko2010adapting} dataset.

\subsubsection{Experiments on PASCAL VOC 2007}
\noindent \textbf{Train/Test Set Acquisition } As a training set, we generate 1080 images with DGCAN from the W-UG subset of CAD-synthetic dataset~\cite{peng2015learning}. These images are equally distributed into 20 PASCAL categories. For evaluation, we crop 14976 patches from 4952 images in the test subset of PASCAL VOC 2007 dataset~\cite{pascal-voc-2007}. The patches are cropped using annotated object bounding boxes and each patch contains only one object. 

\vspace{0.1cm}
\noindent \textbf{Experimental Setup} We evaluate the effectiveness of our approach by training three off-the-shelf DCNN classifiers, \ie ``AlexNet''~\cite{alexnet}, ``VGG-16''~\cite{vgg} and ``ResNet-50'' (\textit{residual net} with 50 layers)~\cite{he2015deep}. In the training process, the networks are initialized with the parameters pre-trained on ImageNet~\cite{ImageNet}. We replace the last output layer with a 20-way classifier and randomly initialize it with $\mathcal{N}(0, 0.01)$. We use mini-batch stochastic gradient descent (SGD) with a momentum of 0.9 to finetune all the layers. The base learning rate is $10^{-3}$ and the weight decay is $5 \times 10^{-4}$. Specifically, we set dropout ratios for \emph{fc6} and \emph{fc7} of ``AlexNet'' to 0.5. We report the results when the training iteration reaches 40k.

\noindent \textbf{Baselines} We compare our approach with the CAD-synthetic method~\cite{peng2015learning}, style transfer~\cite{gatys2015neural} and domain adaptation algorithms~\cite{sun2015return, SunS16a, fernando-iccv13, gong2012geodesic, long2015learning, long2016unsupervised}. 

To compare with state-of-the-art domain adaptation methods, we use the following baselines. \textbf{\textit{CORAL}}~\cite{sun2015return} aligns the feature distribution of source domain ($P^s(\mathbf{x^s}, y^s)$) to target domain ($P^t(\mathbf{x^t}, y^t))$.\textbf{ \textit{DCORAL}} (Deep CORAL)~\cite{SunS16a} incorporates CORAL as a loss layer in the DCNN. \textbf{\textit{SA}} (Subspace Alignment)~\cite{fernando-iccv13} proposes a mapping function to align the subspace of the source domain with the target domain. The subspace is described by the eigenvectors of features~\cite{fernando-iccv13}. \textbf{\textit{GFK}} (Geodesic Flow Kernel)~\cite{gong2012geodesic} models domain discrepancy by integrating numerous subspaces which characterize changes in geometric and statistical properties. Based on these subspaces, a geodesic curve is constructed, a geodesic flow kernel is computed and a kernel-based classifier is trained. \textbf{\textit{DAN}} (Deep Adaptation Network)~\cite{long2015learning} and \textbf{\textit{RTN}} (Residual Transfer Network)~\cite{long2016unsupervised} train deep models with the Maximum Mean Discrepancy~\cite{sejdinovic2013equivalence} loss to align the feature distribution of two domains. 

For equal comparison, we take the same 1080 \textbf{W-UG} images which we utilized to generate our DGCAN-synthetic dataset as the source domain for domain adaptation algorithms. For style-transfer method~\cite{gatys2015neural}, we use the same CAD-synthetic (\textit{content}) images and real images (\textit{style}) to generate new dataset. For \textit{\textbf{SA}}~\cite{fernando-iccv13} and \textit{\textbf{GFK}}~\cite{gong2012geodesic}, we first extract deep features and then apply their model to get the baseline results. For all the baselines, we use the code and experimental settings provided by the authors to run all the experiments. 

\vspace{0.05cm}
\noindent \textbf{Results and Analysis}
The per-category accuracies of AlexNet classifier are presented in Table~\ref{tab_cls}, demonstrating that our approach outperforms competing methods. After applying our approach to CAD-synthetic data, the overall accuracy rises from 18.48\% to 27.46\%.
Additionally, Table~\ref{tab_cls} shows that our approach gains a clear advantage over the state-of-the-art domain adaptation algorithms~\cite{sun2015return, SunS16a, fernando-iccv13, gong2012geodesic, long2015learning, long2016unsupervised} and the style-transfer baseline~\cite{gatys2015neural}. The latter result reveals that aligning the covariance matrix works better than aligning the Gram matrix in the synthetic-to-real domain adaptation scenario. 
In Table~\ref{tab_cls_vgg_res}, we further show VGG and ResNet classifiers trained on the adapted dataset outperform the CAD-synthetic method~\cite{peng2015learning} and domain adaptation methods~\cite{sun2015return, SunS16a, fernando-iccv13, gong2012geodesic}. With our model, the accuracies of VGG and ResNet classifiers rise from 10.3\% to 22.92\% and from 13.13\% to 20.59\%, respectively. We notice that AlexNet achieves the best overall performance. Given that VGG and ResNet have more parameters than AlexNet, we assume that they are overfitting to the generated synthetic dataset, which causes poor generalization to real-image domain.

We visualize how the inter-class confusion mode and the feature embeddings have changed after applying our model, as shown in Figure~\ref{fig_conf}. The confusion matrices on the left show that AlexNet~\cite{alexnet} trained on the CAD-synthetic dataset ($\mathcal{I}_s$) tends to mistake other categories for ``boat'' and ``train''. This phenomenon disappears after applying our model to the CAD-synthetic dataset, as illustrated by the second subplot on the left of Figure~\ref{fig_conf}. This effect is partially explained by the texture synthesizing ability of DGCAN, which provides additional discriminative cues to the CAD-synthetic images. At the feature level, we visualize layer \textit{fc7}'s feature embeddings by t-SNE~\cite{maaten2008visualizing} before and after applying our model, as illustrated in the right two subplots of Figure~\ref{fig_conf}. The t-SNE~\cite{maaten2008visualizing} visualization results clearly show that the features of realistic and synthetic images are better aligned after applying our model to CAD-synthetic images.

\subsubsection{Experiments on the Office Dataset}

We also evaluate our method on the Office benchmark~\cite{saenko2010adapting}, which was introduced specifically for studying the effect of domain shift on object classification.
We evaluate the domain generalization ability of our approach by adapting the CAD domain to the real-image \textit{Amazon} domain (images downloaded from amazon.com) in the Office dataset.

\begin{table*}[t]
\scriptsize
\begin{center}
\begin{tabular}
{c  ||p{0.03cm} p{0.03cm} p{0.03cm} p{0.03cm} p{0.03cm} p{0.03cm} p{0.03cm} p{0.03cm}  p{0.03cm} p{0.03cm} p{0.03cm} p{0.03cm} p{0.03cm} p{0.03cm} p{0.03cm} p{0.03cm} p{0.03cm} p{0.03cm} p{0.03cm} p{0.03cm} p{0.03cm} p{0.03cm} p{0.03cm} p{0.03cm} p{0.03cm} p{0.03cm} p{0.03cm} p{0.03cm} p{0.03cm} p{0.03cm} p{0.16cm}   |c}
\hline
Method&bp&bk&bh&bc&bt&ca&dc&dl&dp&fc&hp&kb&lc&lt&mp&mt&ms&mg&pn&pe&ph&pr&pj&pn&rb&rl&sc&sp&st&td&tc&all\\
\hline
AlexNet-web~\cite{alexnet}&76 &96 &90 &48 &33 &86 &82 &59 &3 &51 &60 &70 &76 &14 &14 &60 &\textbf{86} &32 &32 &\textbf{55 }&42 &52 &30 &12 &24 &43 &41 &16 &\textbf{31} &\textbf{39} &14 &47.30\\
SA-web~\cite{fernando-iccv13}&77 &96 &87 &27 &34 &77 &84 &35 &5 &40 &63 &71 &73 &8 &29 &47 &77 &33 &32 &42 &35 &57 &67 &14 &26 &\textbf{50} &37 &13 &13 &34 &17 &47.18\\
GFK-web~\cite{gong2012geodesic}&10 &94 &85 &17 &11 &73 &76 &26 &15 &10 &65 &72 &\textbf{79} &9 &5 &21 &44 &27 &27 &46 &23 &32 &21 &13 &24 &48 &37 &31 &10 &24 &25 &35.25\\
CORAL-web~\cite{sun2015return}&82 &95 &93 &38 &39 &78 &88 &45 &12 &35 &74 &79 &73 &10 &36 &51 &77 &33 &44 &44 &37 &60 &62 &17 &27 &45 &42 &17 &24 &45 &19 &48.43\\
DAN-web~\cite{long2015learning}&87 &96 &82 &23 &\textbf{58} &87 &90 &35 &2 &36 &78 &85 &74 &9 &57 &\textbf{89} &83 &68 &43 &48 &35 &52 &44 &17 &\textbf{37 }&43 &42 &28 &19 &29 &16 & 49.63\\
\hline
\hline
AlexNet-CAD~\cite{alexnet}&61 &94 &15 &50 &47 &78 &91 &49 &16 &7 &57 &\textbf{88} &72 &26 &70 &71 &49 &73 &32 &20 &\textbf{53} &\textbf{93 }&55 &12 &3 &16 &33 &6 &2 &15 &6 &44.69\\
SA-CAD~\cite{fernando-iccv13}& 78 &94 &83 &52 &42 &84 &85 &38 &5 &46 &65 &78 &62 &10 &41 &49 &72 &30 &32 &36 &40 &60 &50 &16 &25 &34 &40 &10 &6 &27 &21 &47.32\\
GFK-CAD~\cite{gong2012geodesic}&60 &96 &78 &17 &19 &69 &84 &22 &18 &16 &49 &56 &68 &12 &1 &23 &35 &22 &20 &45 &30 &12 &16 &14 &24 &35 &42 &8 &6 &21 &23 &31.38\\
CORAL-CAD~\cite{sun2015return}&60 &98 &\textbf{92} &63 &39 &79 &90 &42 &15 &36 &\textbf{80} &80 &74 &12 &37 &55 &63 &33 &32 &47 &46 &58 &60 &\textbf{19} &30 &39 &\textbf{44} &20 &16 &38 &19 &47.93\\
DAN-CAD~\cite{long2015learning}&90 &96 &62 &62 &35 &76 &86 &57 &26 &24 &66 &82 &72 &31 &70 &72 &66 &67 &46 &30 &26 &52 &63 &11 &13 &16 &32 &38 &18 &16 &20 & 49.27\\
\textbf{DGCAN}&\textbf{91} &\textbf{98} &26 &\textbf{72} &39 &\textbf{89} &\textbf{91} &47 &20 &\textbf{56 }&66 &85 &\textbf{67} &23 &59 &67 &54 &72 &30 &18 &52 &76 &60 &2 &2 &17 &27 &\textbf{42} &23 &22 &14 &\textbf{49.91}\\
\textbf{DGCAN+DAN} & 90 &96 &35 &67 &50 &80 &85 &\textbf{59} &\textbf{27} &48 &77 &58 &74 &\textbf{32} &\textbf{74} &73 &64 &\textbf{80} &\textbf{53} &22 &27 &49 &67 &4 &21 &39 &33 &37 &29 &28 &\textbf{25} &\textbf{51.93}\\
\hline
\end{tabular}
\end{center}
\caption{\textbf{Results on Office Dataset.} We apply \textit{DGCAN} to 775 CAD-synthetic images and train AlexNet classifiers~\cite{alexnet}. The test images come from \textit{Amazon} domain of Office dataset~\cite{saenko2010adapting}. The results clearly shows our approach outperforms the competing baselines~\cite{sun2015return, SunS16a, fernando-iccv13, gong2012geodesic, long2015learning, long2016unsupervised}. The suffix ``-web'' and ``-CAD'' represent the methods are trained on \textit{Webcam} domain~\cite{saenko2010adapting} and CAD-synthetic domain, respectively.}
\label{tab_officeresult}
\vspace{-0.1in}
\end{table*}

\begin{table*}[t]
\small
\begin{center}
\begin{tabular}
{ p{0.5cm} p{1.4cm} p{1.5cm} p{1.4cm} p{1.4cm} p{1.4cm} p{1.4cm} p{1.4cm} p{1.4cm} p{1.4cm} p{1.4cm} p{1.4cm} }

&\includegraphics[width=0.99\linewidth]{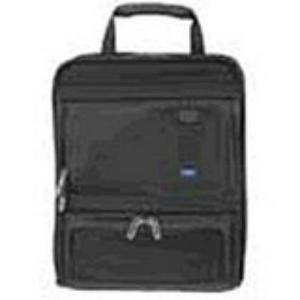}
&\includegraphics[width=0.99\linewidth]{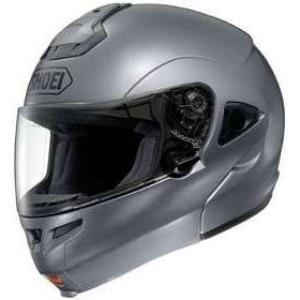}
&\includegraphics[width=0.99\linewidth]{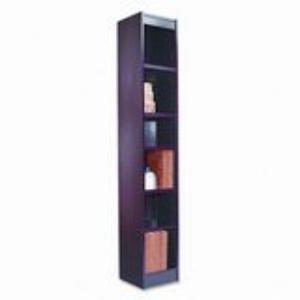}
&\includegraphics[width=0.99\linewidth]{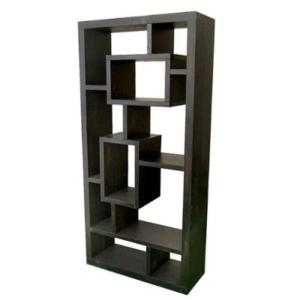}
&\includegraphics[width=0.99\linewidth]{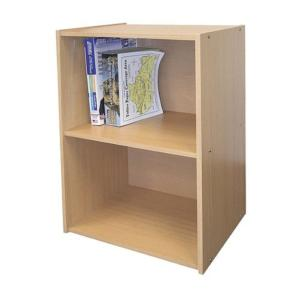}
&\includegraphics[width=0.99\linewidth]{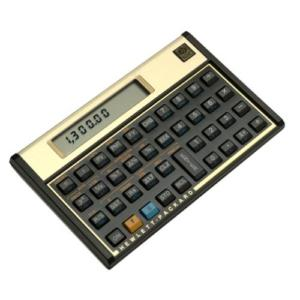}
&\includegraphics[width=0.99\linewidth]{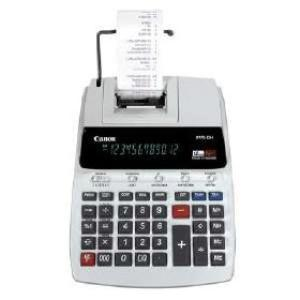}
&\includegraphics[width=0.99\linewidth]{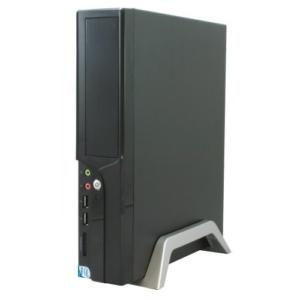}
&\includegraphics[width=0.95\linewidth]{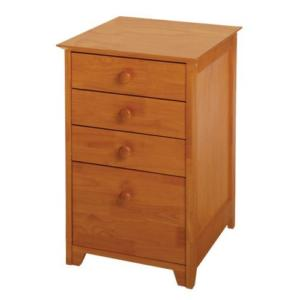}\\
\hline
\textbf{GT} &back pack&bike helmet&bookcase&bookcase&bookcase&calculator&calculator&computer&file cabinet\\

\textbf{Cad} &\textcolor{red}{printer} & \textcolor{red}{printer}& \textcolor{red}{bottle} &\textcolor{red}{computer} &\textcolor{red}{file cabinet} &\textcolor{red}{keyboard} & \textcolor{red}{phone} & \textcolor{red}{printer} & \textcolor{red} {bookcase} \\

\textbf{Ours} & \textcolor{green}{back pack} &\textcolor{green}{bike helmet} &\textcolor{green}{bookcase} &\textcolor{green}{bookcase} &\textcolor{green}{bookcase} &\textcolor{green}{calculator} &\textcolor{green}{calculator} &\textcolor{green}{computer} &\textcolor{green}{file cabinet} \\
\hline

&\includegraphics[width=0.99\linewidth]{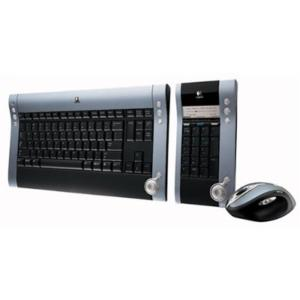}
&\includegraphics[width=0.99\linewidth]{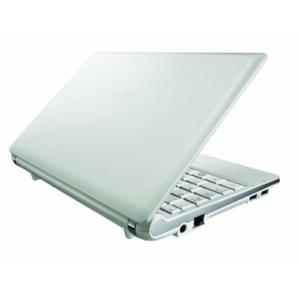}
&\includegraphics[width=0.99\linewidth]{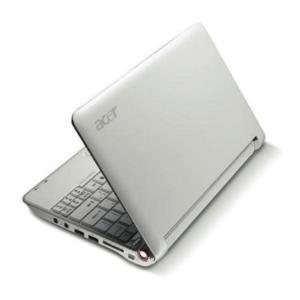}
&\includegraphics[width=0.99\linewidth]{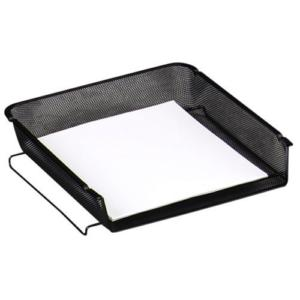}
&\includegraphics[width=0.99\linewidth]{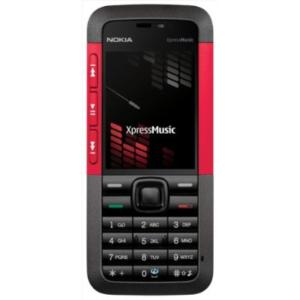}

&\includegraphics[width=0.99\linewidth]{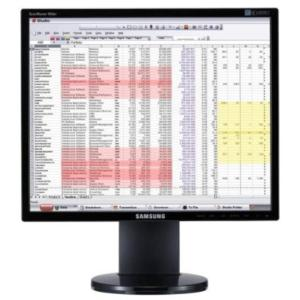}
&\includegraphics[width=0.99\linewidth]{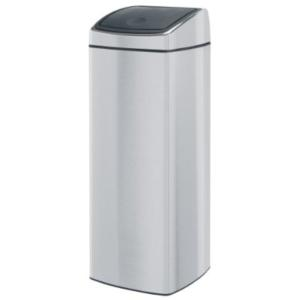}
&\includegraphics[width=0.99\linewidth]{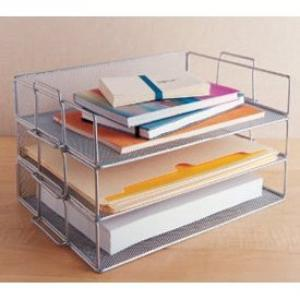}
&\includegraphics[width=0.95\linewidth]{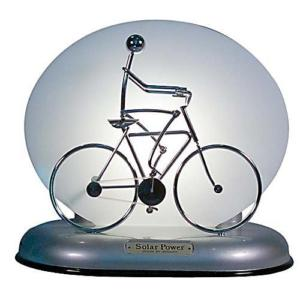}\\       
\hline
\textbf{GT} &keyboard&laptop&laptop&letter tray&mb phone&monitor&trash can&letter tray&desk lamp\\

\textbf{Cad} &\textcolor{red}{computer} & \textcolor{red}{keyboard}& \textcolor{red}{notebook} &\textcolor{red}{monitor} &\textcolor{red}{calculator} &\textcolor{green}{monitor} & \textcolor{green}{trash can} & \textcolor{red}{printer} & \textcolor{red} {bike} \\

\textbf{Ours} & \textcolor{green}{keyboard} &\textcolor{green}{laptop} &\textcolor{green}{laptop} &\textcolor{green}{letter tray} &\textcolor{green}{mb phone} &\textcolor{red}{keyboard} &\textcolor{red}{ring binder} &\textcolor{red}{punchers} &\textcolor{red}{bike} \\
\hline

\end{tabular}
\end{center} 
\caption{Instances from the Amazon domain of the Office dataset and the corresponding labels predicted by the baseline trained on CAD-synthetic images (CAD), and our model. We show examples where our model improves on the baseline, as well as typical failure cases.}
\label{tab_cad_dgcan}
\vspace{-0.2in}
\end{table*}

\vspace{0.1cm}
\noindent \textbf{Train/Test Set Acquisition} 
We apply our model to the 775 CAD-synthetic images provided by~\cite{peng2015learning} to generate the training dataset. These CAD-synthetic images are rendered to train object detectors for Office dataset. To collect $\mathcal{I}_t$ (natural images), for each category, we downloaded 4$\sim$5 images from Google by searching the category's name. The test set comes from the Office Dataset~\cite{saenko2010adapting}, which has the same 31 categories (\eg backpack, cups, \etc) in three domains, \ie \textit{Amazon}, \textit{Webcam} (collected by webcam) and \textit{DSLR} (collected by DSLR camera). Specifically, we use the \textit{Amazon} set (2817 images) as the test set in our experiments as it is the most challenging setting, and because this domain significantly differs from PASCAL (see Table~\ref{tab_cad_dgcan} for examples).

\vspace{0.1cm}
\noindent \textbf{Baselines} We compare our approach to two sets of baselines, with one set trained on another real image domain in Office (\textit{Webcam} domain, 795 images) and the other trained on the CAD-synthetic domain (775 images). In both sets, we compare  to the basic AlexNet~\cite{alexnet} model (no adaptation) and domain adaptation algorithms~\cite{fernando-iccv13, gong2012geodesic, sun2015return, long2015learning}.

\vspace{0.1cm}
\noindent \textbf{Results} The results demonstrate that our approach performs strongly on this benchmark, as can be seen in Table~\ref{tab_officeresult}. The overall classification accuracy of our model is 49.91\% versus 44.69\% for a classifier trained on the CAD-synthetic domain directly. The table also shows that \textit{DGCAN} beats other baselines~\cite{fernando-iccv13, gong2012geodesic, sun2015return} and classifiers trained on real images (\textit{Webcam} domain), and is slightly better than the domain alignment network (DAN) of~\cite{long2015learning}. We note here that this and other unsupervised domain adaptation baselines make use of the test data to train the alignment models (transductive training). On the other hand, our method did not use the test images for training, but performs well nonetheless.

We further show that our model is complementary with transductive domain adaptation algorithms. We set the newly generated dataset as the new source domain and adapt it to the real Amazon domain with \textit{\textbf{DAN}}~\cite{long2015learning}. As showed in Table~\ref{tab_officeresult}, this boosts the performance from 49.63\% (49.27\%) for \textit{DAN} trained on real (CAD-synthetic) domain to 51.93\%.

Table~\ref{tab_cad_dgcan} shows some results for which the classifier trained on CAD-synthetic images fails to predict the correct labels while our model predicts the right ones, as well as some representative mistakes. The results show the potential to generate better training data for a large variety of object categories. An interesting example is the ``desk lamp'' with a toy bike in the middle, causing both models to mistake it for a bike.

\section{Conclusion}
Generating large-scale training examples from 3D CAD models is a promising alternative to expensive data annotation. However,  the domain discrepancy between CAD-synthetic images and real images severely undermines the performance of deep learning models on real world applications.

In this work, we have proposed and implemented a Deep Generative Correlation Alignment Network to adapt the CAD-synthetic domain to realistic domains by generating images with more natural feature statistics. We demonstrated that leveraging $\ell^2$ loss to preserve the content of the CAD models in feature space and applying the second-order \emph{CORAL} loss to diminish the domain discrepancy are effective in synthesizing adapted training images. We empirically and experimentally show that DGCAN-synthetic images are more suitable for training deep CNNs than CAD-synthetic ones. An extensive evaluation on standard benchmarks demonstrates the feasibility and effectiveness of the proposed approach against previous methods. We believe our model can be generalized to other generic tasks such as pose estimation, saliency detection and robotic grasping.

{\small
\bibliographystyle{ieee}
\bibliography{egbib}
}

\end{document}